\documentclass[10pt,twocolumn,letterpaper]{article}

\usepackage{cvpr}
\usepackage{times}
\usepackage{epsfig}
\usepackage{graphicx}
\usepackage{amsmath}
\usepackage{amssymb}
\usepackage{algorithm}
\usepackage{algpseudocode}
\usepackage{wrapfig}
\usepackage{algpseudocode}

\usepackage[breaklinks=true,bookmarks=false]{hyperref}

\cvprfinalcopy 

\def\cvprPaperID{****} 

\ifcvprfinal\fi
\begin{document}

\title{Expert Gate: Lifelong Learning with a Network of Experts}
\author{Rahaf Aljundi
\and
Punarjay Chakravarty
\\
{ \footnotesize $\{$rahaf.aljundi, Punarjay.Chakravarty, Tinne.Tuytelaars$\}$@esat.kuleuven.be}
\\
KU Leuven, ESAT-PSI, IMEC, Belgium \\
\and
Tinne Tuytelaars
}

\maketitle

\begin{abstract}
In this paper we introduce a model of lifelong learning, based on a Network of Experts. 
New tasks / experts are learned and added to the model sequentially, building on what was
learned before. To ensure scalability of this process, data from previous tasks 
cannot be stored and hence is not available when learning a new task. 
A critical issue in such context, not addressed
in the literature so far, relates to the decision which expert to deploy at test time.
We introduce a set of gating autoencoders that learn a representation for the task at hand, and, at test time, automatically forward the test sample to the relevant expert.
This also brings memory efficiency as only one expert network has to be loaded into memory at any given time.
Further, the autoencoders inherently capture the relatedness of one task to another, 
based on which the most relevant prior model to be used for training a new expert, with fine-tuning or learning-without-forgetting, can be selected.
We evaluate our method on image classification and video prediction problems. 

\end{abstract}

\section{Introduction}
\label{sec:Into}


In the age of deep learning and big data, we face a situation
where we train ever more complicated models with ever increasing amounts of data. We have different models for different tasks trained on different datasets, each of which is an expert on its own domain, but not on others. 
In a typical setting, each new task comes with its own dataset.
Learning a new task, say scene classification based on a pre-existing object recognition network trained on ImageNet, requires adapting the model to the new set of classes and
fine-tuning it with the new data. The newly trained network performs well on the new task, but has a degraded performance on the old ones. This is called {\em catastrophic forgetting}~\cite{goodfellow2013empirical}, and is a major problem facing life long learning techniques \cite{silver2002task,silver2013lifelong,rusu2016progressive}, where new tasks and datasets are added in a sequential manner.
\begin{figure}[t]
\centering
\includegraphics[width=0.4\textwidth]{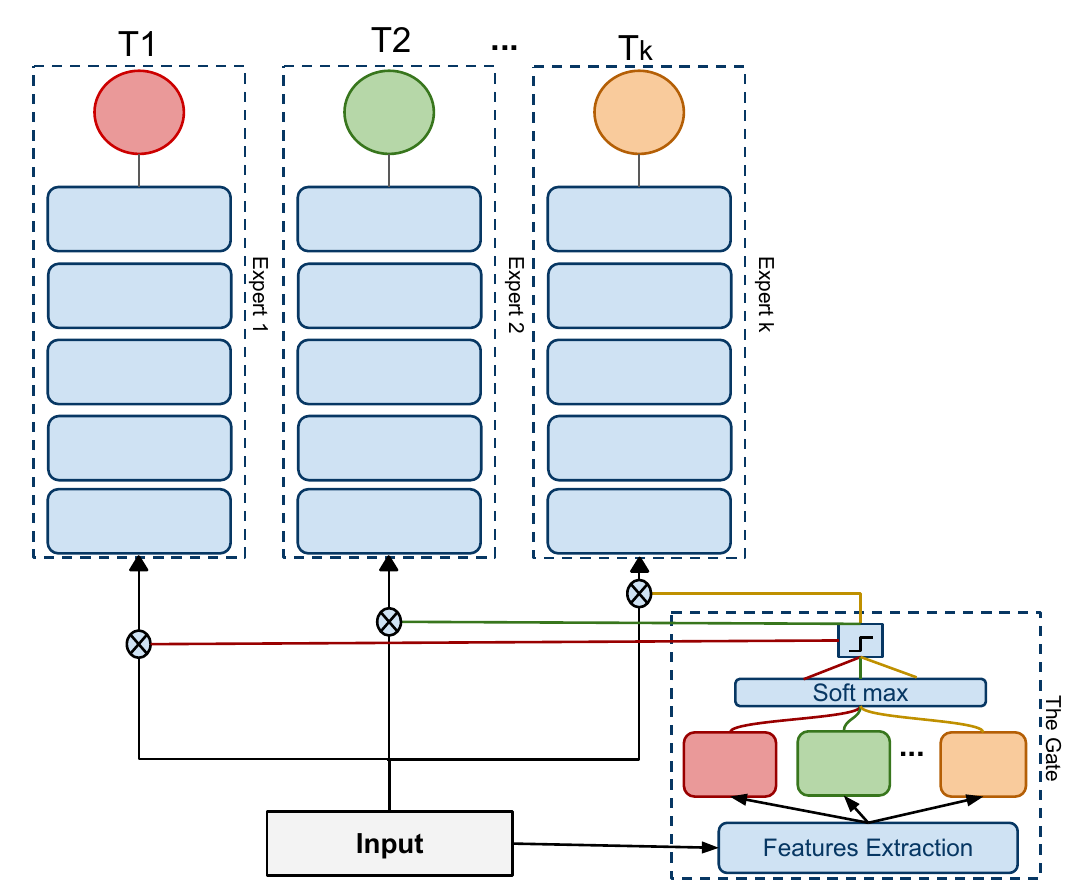}
  \caption{The architecture of our Expert Gate system.}
   
  \label{fig:system}
  \vspace*{-0.5cm} 
\end{figure}


Ideally, a system should be able to operate on different tasks and domains and give the best performance on each of them. For example, an image classification system that is able to operate on generic as well as fine-grained classes, and in addition performs action and scene classification. If all previous training data were available, a direct solution would be to jointly train a model on all the different tasks or domains. Each time a new task arrives along with its own training data, new layers/neurons are added, if needed, and the model is retrained on all the tasks.
Such a solution has three main drawbacks. The first is the risk of the negative inductive bias when the tasks are not related or simply adversarial. Second, a shared model might fail to capture specialist information for particular tasks as joint training will encourage a hidden representation beneficial for all tasks. Third, each time a new task is to be learned, the whole network needs to be re-trained.
Apart from the above drawbacks, the biggest constraint with joint training is that of keeping all the data from the previous tasks. This is a difficult requirement to be met, especially in the era of big data.
%
For example, ILSVRC~\cite{ILSVRC15} has 1000 classes, with over a million images, amounting to 200 GB of data. Yet the AlexNet
model trained on the same dataset, is only 200 MB, a difference in size of three orders of magnitude. With increasing amounts of data collected, it becomes less and less feasible to store all the training data, and more practical to just store the models learned from the data.

Without storing the data, one can consider strategies like using the previous model to generate virtual samples (i.e. use the soft outputs of the old model on new task data to generate virtual labels) and use them in the retraining phase \cite{caruana1998multitask,li2016learning,silver2002task}. 
This works to some extent, but is unlikely to scale as repeating this scheme a number of times
causes a bias towards the new tasks and an exponential buildup of errors on the older ones,
as we show in our experiments.
Moreover, it suffers from the same drawbacks as the joint training described above. 
Instead of having a  network that is jack of all trades and master of none, we stress the need for having different specialist or expert models for different tasks, as also advocated
in~\cite{hinton2015distilling,jacobs1991adaptive,rusu2016progressive}. 
Therefore we build a Network of Experts, where a new expert model is added whenever a new task
arrives 
and knowledge is transferred from previous models. 

With an increasing number of task specializations, the number of expert models increases. Modern GPUs, used to speed up training and testing of neural nets, have limited memory (compared to CPUs), and can only
load a relatively small number of models at a time. We obviate the need for loading all the models by
learning a gating mechanism that uses the test sample to decide which expert to activate ( see Figure \ref{fig:system} ).
For this reason, we call our method {\em Expert Gate}.

Unlike \cite{kokkinos2016ubernet}, who train one Uber network for performing vision tasks as diverse as semantic segmentation, object detection and human body part detection, our work focuses on tasks with a similar objective.
For example, imagine a drone trained to fly through an environment using its frontal camera. For optimal performance, it needs to deploy different models for different environments such as indoor, outdoor or forest. 
Our gating mechanism then selects a model on the fly based on the input video.
Another application could be a visual question answering system, that has multiple models trained using images from different domains.
Here too, our gating mechanism could use the data itself to select the associated task model. 

Even if we could deploy all the models simultaneously, 
selecting the right expert model 
is not straightforward.
Just using the output of the highest scoring expert is no guarantee for success
as neural networks can erroneously give high
confidence scores, as shown in \cite{nguyen2015deep}.  
We also demonstrate this in our experiments.
Training a discriminative classifier to distinguish between tasks
is also not an option since that would again require storing all training data. 
%
What we need is a task recognizer that can tell the relevance of its associated task model for a given test sample. This is exactly what our gating mechanism provides. In fact, also the prefrontal cortex of the primate brain is considered to have neural representations of task context that act as a gating in different brain functions~\cite{mante2013context}. 


We propose to implement such task recognizer  using an undercomplete autoencoder as a gating mechanism.
We learn for each new task or domain, a gating function that captures the shared characteristics among the training samples and can recognize similar samples at test time. We do so using a one layer under-complete autoencoder. Each autoencoder is trained along with the corresponding expert model and maps the training data to its own lower dimensional subspace. At test time, each task autoencoder projects the sample to its learned subspace and measures the reconstruction error due to the projection. The autoencoder with the lowest reconstruction error is used like a switch, selecting the corresponding expert model (see Figure \ref{fig:system}).

Interestingly, such autoencoders can 
also be used to evaluate {\em task relatedness} at training time, which in turn can be used  to determine which prior model is more relevant to a new task. 
We show how, based on this information, Expert Gate can decide which specialist model to transfer knowledge from when learning a new task and whether to use fine-tuning or learning-without-forgetting~\cite{li2016learning}.

To summarize, our contributions are the following.
We develop Expert Gate, a lifelong learning system that can sequentially deal with new tasks without storing all previous data. It automatically selects the most related prior task to aid learning of the new task. At test time, the appropriate model is loaded automatically to deal with the task at hand.
We evaluate our gating network on image classification and video prediction problems. 

The rest of the paper is organized as follows. We discuss related work in Section \ref{relWork}. Expert Gate is detailed in Section~\ref{method}, followed by experiments 
in Section \ref{sec:expres}. We finish with concluding remarks and future work in Section \ref{conclusions}.

\section{Related Work}
\label{relWork}
\textbf{Multi-task learning}
Our end goal is to develop a system that can reach expert level performance on multiple tasks, with tasks learned sequentially. As such, it lies at the intersection between multi-task learning and lifelong learning. 
Standard multi-task learning \cite{caruana1998multitask} aims at learning multiple tasks in a joint manner. 
The objective is to use knowledge from different tasks, the so called inductive bias \cite{mitchell1980need}, in order to improve performance on individual tasks.
Often one shared model is used for all tasks. This has the benifit of relaxing the number of required samples per task but could lead to suboptimal performance on the individual tasks.
On the other hand, multiple models can be learned, that are each optimal for their own task, but utilize inductive bias / knowledge from other models \cite{caruana1998multitask}.

To determine which related tasks to utilize,  ~\cite{thrun1998clustering}  cluster the  tasks  based on the mutual information gain when using the information from one task while learning another. 
This is an exhaustive process. As an alternative, 
\cite{jacob2009clustered,xue2007multi,kumar2012learning}  assume that the parameters of related task models lie close by in the original space or in a lower dimensional subspace and thus cluster the tasks' parameters. They first learn task models independently, then use the tasks within the same cluster to help improving or relearning their models. 
This requires learning individual task models first.
Alternatively, we use our tasks autoencoders, that are fast to train, to identify related tasks.

\textbf{Multiple models for multiple tasks}
One of the first examples of using multiple models, each one handling a subset of tasks, was by Jacobs et al.~\cite{jacobs1991adaptive}. They trained an adaptive mixture of experts
(each a neural network) for multi-speaker vowel recognition and used a separate gating network to determine which network to use for each sample.
They showed that this setup
outperformed a single shared model. A downside, however, was that each training sample needed to pass through each  expert, for the gating function to be learned.
To avoid this issue, 
a mixture of one generalist model and many specialist models has been proposed~\cite{ahmed2016network,hinton2015distilling}. 
%
At test time, the generalist model acts as a gate, forwarding the sample to the correct network. 
However, unlike our model, these approaches require all the data to be available for learning the generalist model, which needs to be retrained each time a new task arrives. 

\textbf{Lifelong learning without catastrophic forgetting} 
In sequential lifelong learning, knowledge from previous tasks is leveraged to improve the training of new tasks, while taking care not to forget old tasks, i.e. preventing catastrophic forgetting \cite{goodfellow2013empirical}. 
Our system obviates the need for storing all the training data collected during the lifetime of an agent, by learning task autoencoders that learn the distribution of the task data, and hence, also capture the meta-knowledge of the task. This is one of the desired characteristics of a lifelong learning system, as outlined by Silver et al.~\cite{silver2013lifelong}.
The constraint of not storing all previous training data has been looked at previously by Silver and Mercer~\cite{silver2002task}. They use the output of the previous task networks given new training data, called virtual samples, to regularize the training of the networks for new tasks. This improves the new task performance by using the knowledge of the previous tasks. More recently, the Learning without Forgetting framework of~\cite{li2016learning} uses a similar regularization strategy, but learns a single network for all tasks: they finetune a previously trained network (with additional task outputs) for new tasks. The contribution of previous tasks/networks in the training of new networks is determined by task relatedness metrics in \cite{silver2002task}, while in \cite{li2016learning}, 
all previous knowledge  is used,
regardless of task relatedness.
\cite{li2016learning} demonstrates sequential training of a network for only two tasks. In our experiments, we show that the shared model gets worse when extended to more than two tasks, especially when task relatedness is low.

Like us, two recent architectures, namely the progressive network \cite{rusu2016progressive} and the modular block network \cite{terekhov2015knowledge}, 
also use multiple networks, a new one for each new task. They add new networks as additional columns with lateral connections to the previous nets. These lateral connections mean that each layer in the new network is connected to not only its previous layer in the same column, but also to previous layers from all previous columns. This allows the networks to transfer knowledge from older to newer tasks. However, in these works, choosing which column to use for a particular task at test time is done manually, and the authors leave its automation as future work. 
Here, we propose to use an autoencoder to determine which model, and consequently column, is to be selected for a particular test sample. 

\section{Our Method}
\label{method}
We consider the case of lifelong learning or sequential learning where tasks and their corresponding data come one after another. For each task, we learn
a specialized model (expert) by transferring knowledge from previous tasks -- in particular, we build on the {\em most related} previous task. Simultaneously we learn a gating function that captures the characteristics of each task.
This gate forwards the test data to the corresponding expert resulting in a high performance over all learned tasks. 

%
The question then is: how to learn such a gate function to differentiate between tasks, without having access to the training data of previous tasks? To this end, we learn a low dimensional subspace for each task/domain. At test time we then select the representation (subspace) that best fits the test sample. We do that using an undercomplete autoencoder per task. 
Below, we first describe this autoencoder in more detail (Section~\ref{sec:autoencoder}). Next, we explain how to use them for selecting the most relevant expert (Section~\ref{sec:selection}) and for estimating task relatedness (Section~\ref{sec:relatedness}).

  \subsection{The Autoencoder Gate}
  \label{sec:autoencoder}
An autoencoder \cite{bourlard1988auto} is a neural network that learns to produce an output similar to its input \cite{Goodfellow-et-al-2016-Book}. The network is composed of two parts, an encoder $f=h(x)$, which maps the input $x$ to a code $h(x)$ and a decoder $r=g(h(x))$, that maps the code to a reconstruction of the input. 
The loss function $L(x,g(h(x)))$ is simply the reconstruction error. 
The encoder learns, through a hidden layer, a lower dimensional representation (undercomplete autoencoder)  or a higher dimensional representation (overcomplete autoencoder) of the input data, guided by regularization criteria to prevent the autoencoder from copying its input.  
A linear autoencoder with a Euclidean loss function learns the same subspace as PCA. 
However, autoencoders with non-linear functions yield better dimensionality reduction compared to PCA~ \cite{hinton2006reducing}.
This motivates our choice for this model.

Autoencoders are usually used to learn feature representations in an unsupervised manner or for  dimensionality reduction. Here, we use them for a different goal.
The lower dimensional subspace learned by one of our undercomplete autoencoders will be maximally sensitive to variations observed in the task data but insensitive to changes orthogonal to the manifold. In other words, it represents only the variations that are needed to reconstruct relevant samples.
%
\begin{figure}[t]
\centering
\includegraphics[width=0.35\textwidth,height=0.18\textheight]{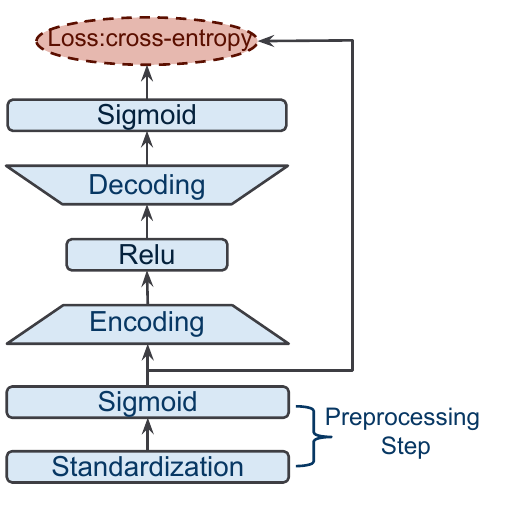}
  \caption{Our autoencoder gate structure.}    
  \label{fig:structure}
  \vspace*{-0.5cm} 
  \end{figure}
Our main hypothesis is that the autoencoder of one domain/task should thus be better at reconstructing the data of that task than the other autoencoders. Comparing the reconstruction errors of the different tasks' autoencoders then allows to successfully forward a test sample to the most relevant expert network. 
It has been stated by~\cite{alain2014regularized} that in regularized (over-complete) autoencoders, the opposing forces between the risk and the regularization term result in a score like behavior for the reconstruction error. As a result, a zero reconstruction loss means a zero derivative which could be a local minimum or a local maximum. However, we use an unregularized one-layer under-complete autoencoder and for these, it has been shown~\cite{bengio2009learning,marc2007unified} that the mean squared error criterion we use as reconstruction loss estimates the negative log-likelihood. There is no need in such a one-layer autoencoder to add a regularization term to pull up the energy on unseen data because the narrowness of the code already acts as an implicit regularizer.

\textbf{Preprocessing} 
We start from a robust image representation $x$, namely the activations of the last convolutional layer of AlexNet pretrained on ImageNet. Before the encoding layer, we pass this input through a preprocessing step, where the input data is standardized, followed by a sigmoid function. The standardization of the data, i.e. subtracting the mean and dividing the result by the standard deviation, is essential as it increases the robustness of the hidden representation to input variations. Normally, standardization is done using the statistics of the data that a network is trained on, but in this case, this is not a good strategy. This is because, at test time, we compare the relative reconstruction errors of the different autoencoders.
Different standardization regimes lead to non-comparable reconstruction errors. Instead, we use the statistics of Imagenet for the standardization of each autoencoder. Since this is a large dataset it gives a good approximation of the distribution of natural images.
After standardization, we apply the sigmoid function to map the input to a range of $[0 \, 1]$. 

\textbf{Network architecture} 
We design a simple autoencoder that is no more complex than one layer in a deep model, with a one layer encoder/decoder (see Figure~\ref{fig:structure}). 
The encoding step consists of one fully connected layer followed by ReLU~\cite{zeiler2013rectified}. We make use of ReLU activation units as they are fast and easy to optimize. ReLU also introduces sparsity in the hidden units which leads to better generalization. 
For decoding, we use again one fully connected layer, but now followed by a sigmoid.
The sigmoid yields values between $ [0 \, 1]$, which allows us to use cross entropy as the loss function. At test time, we use the Euclidean distance to compute the reconstruction error.


\subsection{Selecting the most relevant expert}
\label{sec:selection}
At test time, and after learning the autoencoders for the different tasks, we add a softmax layer that takes as input the reconstruction errors $er_i$ from the different tasks autoencoders given a test sample $x$. 
The reconstruction error $er_i$ of the $i$-th autoencoder is the output of the loss function given the input sample $x$.
The softmax layer gives a probability $p_i$ for each task autoencoder indicating its confidence:
\begin{equation}
p_i=\frac{ exp(-er_i/t)}{\sum_j exp (-er_j/t)}
\end{equation}
where $t$ is the temperature. We use a temperature value of 2 as in  \cite{hinton2015distilling,li2016learning} leading to soft probability values. 
Given these confidence values, we load the expert model associated with the most confident autoencoder. For tasks that  have some overlap, it may be convenient to activate more than one expert model instead of taking the max score only.
This can be done by setting a threshold on the confidence values, see section \ref{sec:expres-gate}. 

\subsection{Measuring task relatedness}
 \label{sec:relatedness} 
 Given a new task $T_k$ associated with its data $D_k$, we first learn an autoencoder for this task $A_k$. Let $T_a$ be a previous task with associated autoencoder $A_a$. We want to measure the task relatedness between task $T_k$  and task $T_a$. Since we do not have access to the data of task $T_a$, we use the validation data from the current task $T_k$. We compute the average reconstruction error $Er_k$ on the current task data made by the current task autoencoder $A_k$ and, likewise, the average reconstruction error $Er_a$ made by the previous task autoencoder $A_a$ on the current task data.
 The relatedness between the two tasks is then computed:
 \begin{equation}
 Rel(T_k,T_a)=1-(\frac {Er_a-Er_k}{Er_k})
 \end{equation}
 Note that the relatedness value is not symmetric. Applying this to every previous task, we get a relatedness value to each previous task. 
 
We exploit task relatedness in two ways. First, we use it to select the most related task to be used as prior model for learning the new task. Second, we exploit the level of task relatedness to determine which transfer method to use: fine-tuning or learning-without-forgetting (LwF)~\cite{li2016learning}.
We found in our experiments that LwF only outperforms fine-tuning when the two tasks are sufficiently related.
When this is not the case, 
enforcing the new model to give similar outputs for the old task
may actually hurt performance.
Fine-tuning, on the other hand, only uses the previous task parameters as a starting point and is less sensitive to the level of task relatedness.
Therefore, 
we apply a threshold on the task relatedness value to decide when to use LwF and when to fine-tune.  
Algorithm  \ref{algo} shows the main steps of our Expert Gate in both  training and test phase.
 \begin{algorithm}[t]
\caption{Expert Gate }\label{algo}
  \begin{algorithmic}[1]
  \Statex \textbf{\textit{Training Phase}} input: expert-models $(E_1,.,E_j)$, tasks-autoencoders $(A_1,.,A_j)$, new task ($T_k$), data ($D_k$) ; output: $E_k$
      \State  $A_k$ =train-task-autoencoder $(D_k)$
      \State (\textit{rel,rel-val})=select-most-related-task($D_k$,$A_k$,\{A\})
      
      \If{  \textit{rel-val} $>$\textit{rel-th}}
      \State $E_k$=LwF$(E_{rel},D_k)$
    \Else
    \State $E_k$=fine-tune$(E_{rel},D_k)$
  \EndIf
  \Statex  \textbf{\textit{ Test Phase}}  input: $x$ ; output:  prediction
  \State $i$=select-expert$(\{A\},x)$
  \State  prediction = activate-expert$(E_{i},x)$
 
  \end{algorithmic}
  \end{algorithm}

\section{Experiments}
\label{sec:expres}
First, 
we compare our method against various baselines on a set of three image classification tasks (Section~\ref{sec:expres-baselines}).
Next, we analyze our gate behavior in more detail on a bigger set of tasks (Section~\ref{sec:expres-gate}), followed by an analysis of our task relatedness measure
(Section~\ref{sec:expres-related}).
Finally, we test Expert Gate on a video prediction problem (Section~\ref{sec:expres-video}).

\noindent
\textbf{Implementation details}
We use the activations of the last convolutional layer of an AlexNet  pre-trained with ImageNet as image representation for our autoencoders. 
We experimented with the size of the hidden layer in the autoencoder, trying sizes of 10, 50, 100, 200 and 500, and found an optimal value of 100 neurons. This is a good compromise between complexity and performance. If the task relatedness is higher than $0.85$, we use LwF; otherwise, we use fine-tuning. 
We use the MatConvNet framework \cite{vedaldi2015matconvnet} for all our experiments.

\subsection{Comparison with baselines}
\label{sec:expres-baselines}
We start with the sequential learning of three image classification tasks: in order, we train on MIT \textit{Scenes} \cite{quattoni2009recognizing} for scene classification, Caltech-UCSD \textit{ Birds}~\cite{WelinderEtal2010} for fine-grained bird classification and Oxford \textit{Flowers}~\cite{Nilsback08} for fine-grained flower classification.
To simulate a scenario in which an agent or robot has some prior knowledge, and is then exposed to datasets in a sequential manner, we start off with an AlexNet model pre-trained on ImageNet. We compare against the following baselines:\\
{\em 1. A single jointly-trained model}: Assuming all training data is always available, this model is jointly trained (by finetuning an AlexNet model pretrained on ImageNet) 
for all three tasks together.\\
{\em 2. Multiple fine-tuned models}: Distinct AlexNet models (pretrained on ImageNet) are finetuned separately, one for each task. At test time, an oracle gate is used, i.e. a test sample is always evaluated by the correct model.
\\
{\em 3. Multiple LwF  models}: Distinct models are learned with learning-without-forgetting~\cite{li2016learning}, one model per new task, always using AlexNet pre-trained on ImageNet as previous task. This is again combined with an oracle gate.\\
{\em 4. A single fine-tuned model}: one AlexNet model (pre-trained on ImageNet) sequentially fine-tuned on each task. \\
{\em 5. A single LwF model}: 
LwF sequentially applied to multiple tasks. Each new task is learned with all the outputs of the previous network as soft targets for the new training samples. So, a network (pre-trained on ImageNet) is first trained for Task 1 data without forgetting ImageNet (i.e. using the pretrained AlexNet predictions as soft targets). Then, this network is trained with Task 2 data, now using ImageNet and Task 1 specific layers outputs as soft targets; and so on.

\begin{wrapfigure}{r}{0.2\textwidth}
 \vspace*{-8pt} 
\centering
   \includegraphics[width=0.2\textwidth]{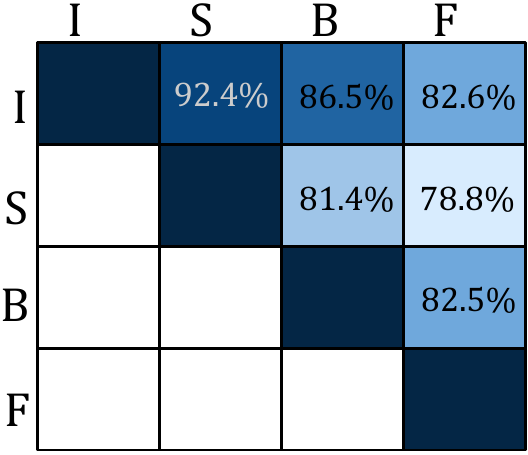}

   \caption{Task relatedness. First letters indicate tasks. } 
   \label{fig:rel3}
 \vspace*{-10pt} 
\end{wrapfigure}

For baselines with multiple models (2 and 3), we rely on an oracle gate to select the right model at test time. So reported numbers for these are upper bounds of what can be achieved in practice. The same holds for baseline 1, as it assumes all previous training data is stored and available.
Table \ref{tab:sequential} shows the  classification accuracy achieved on the test sets of the different tasks.
For our Expert Gate system and for each new task, we first select the most related previous task (including ImageNet) and then learn the new task expert model by transferring knowledge from the most related task model, using LwF or finetuning.
\begin{table}
\footnotesize
\caption{\label{tab:sequential}Classification accuracy for the sequential learning of 3 image classification tasks. Methods with * assume all previous training data is still available, while methods with ** use an oracle gate to select the proper model at test time.}
\centering
\begin{tabular}{ | l | c| c | c ||c |}
\hline
Method & Scenes &Birds & Flowers&avg \\  \hline 
Joint Training* & 63.1 & 58.5 & 85.3& 68.9 \\  \hline  
Multiple fine-tuned models** &  63.4 &  56.8 &85.4&68.5  \\ 

Multiple LwF models** &  63.9 &  58.0 &84.4 &68.7 \\ \hline 

Single fine-tuned model & 63.4 & - & -  &-\\ 
             & 50.3 & 57.3 & - &- \\ 
			 & 46.0 & 43.9 &84.9 &58.2 \\ \hline
Single LwF model &  63.9 &  - &- &- \\ 
            &  61.8 &  53.9 &-&-  \\ 
			&61.2&53.5& 83.8&66.1\\   \hline
Expert Gate (ours) &  63.5 & 57.6 &84.8&68.6  \\ \hline
\end{tabular}
\vspace*{-0.5cm} 
\end{table}
For the \textit{Single fine-tuned model} and \textit{Single LwF model}, we also report intermediate results in the sequential learning. 
When learning multiple models (one per new task), LwF improves over vanilla fine-tuning for Scenes and Birds, as also reported  by \cite{li2016learning}\footnote{Note these numbers are not identical to \cite{li2016learning} but show similar trends. At the time of experimentation, the code for LwF was not available, so we implemented this ourselves in consultation with the authors of~\cite{li2016learning}, and used parameters provided by them.}. However, for Flowers, performance degrades compared to   fine-tuning. 
We measure a lower degree of task relatedness to ImageNet for Flowers than for Birds or Scenes (see Figure \ref{fig:rel3}) which might explain this effect.
Comparing the \textit{Single fine-tuned model} (learned sequentially) with the \textit{Multiple fine-tuned models}, we observe an increasing drop in performance on older tasks: sequentially fine-tuning a single model for new tasks shows catastrophic forgetting and is not a good strategy for lifelong learning.
%
The \textit{Single LwF model} is less sensitive to forgetting on previous tasks.
However, it is still inferior to training exclusive models for those tasks (\textit{Multiple fine-tuned / LwF models}), both for older as well as newer tasks. Lower performance on previous tasks is because 
of a buildup of errors and degradation of the soft targets of the older tasks. This results in LwF failing to compensate for forgetting in a sequence involving more than 2 tasks. This also adds noise in the learning process of the new task. Further, the previous tasks have varying degree of task relatedness. 
On these datasets, we systematically observed the largest task relatedness values for ImageNet (see Figure \ref{fig:rel3}). 
Treating all the tasks equally prevents the new task from getting the same benefit of ImageNet as in the \textit{Multiple LwF models} setting.
Our \textit{Expert Gate} always correctly identifies the most related task, i.e. ImageNet. Based on the relatedness degree, it used LwF for Birds and Scenes, while fine-tuning was used for Flowers. As a result, the best expert models were learned for each task. At test time, our gate mechanism succeeds to select the correct model for $99.2 \%$ of the test samples.
This leads to superior results to those achieved by the other two sequential learning strategies (\textit{Single fine-tuned model} and \textit{Single LwF model}). We achieve comparable performance on average to the \textit{Joint Training} that has access to all the tasks data. Also, performance is on par with  \textit{Multiple fine-tuned models} or \textit{Multiple LwF models} that both assume having the task label for activating the associated model.   
\vspace*{-0.1cm} 
\begin{table*}
\caption{\label{tab:six} Classification accuracy for the sequential learning of 6 tasks. Method with * assumes all the training data is available.}
\centering
\begin{tabular}{ | l | c| c | c | c | c | c ||c|}
\hline
Method & Scenes &Birds & Flowers & Cars & Aircrafts & Actions &avg\\  \hline
Joint Training* & 59.5 & 56.0 & 85.2 &77.4& 73.4 &47.6 &66.5\\  \hline  
Most confident model & 40.4 & 43.0 & 69.2 &78.2&54.2&8.2 &48.7\\  \hline  


Expert Gate  &60.4 &57.0 &84.4 & 80.3 &72.2&49.5&67.3\\ \hline

\end{tabular}
\vspace*{-0.2cm} 
\end{table*}

\subsection{Gate Analysis}
\label{sec:expres-gate}

\begin{table*}

\caption{\label{tab:gate}Results on discriminating between the 6 tasks (classification accuracy)}
\centering
\begin{tabular}{ | l | c| c | c | c | c | c || c|}
\hline
Method & Scenes &Birds & Flowers & Cars & Aircrafts & Actions &avg\\  \hline
Discriminative Task Classifier - \small {\em using all the tasks data} &97.0 & 98.6 & 97.9 & 99.3& 98.8&95.5&97.8\\  \hline 
Expert Gate (ours) - \small {\em no access to the previous tasks data}  &94.6 &97.9& 98.6 & 99.3&97.6&98.1&97.6\\ \hline
\end{tabular}
\vspace*{-0.5cm} 
\end{table*}
The goal of this experiment is to further evaluate our Expert Gate's ability in successfully selecting the relevant network(s) for a given test image. 
For this experiment, we add 3 more tasks: Stanford  \textit{Cars} dataset~\cite{krause20133d} for fine-grained car classification, 
FGVC-\textit{Aircraft} dataset~\cite{maji13fine-grained} for fine-grained classification of aircraft, and
VOC \textit{Actions}, the human  action classification subset of VOC  challenge 2012~\cite{pascal-voc-2012}. 
This last dataset has multi-label annotations. For sake of consistency, we only use the actions with single label. For these newly added datasets, we use the bounding boxes instead of the full images as the images might contain more than one object. So in total we deal with 6 different tasks: Scenes, Birds, Flowers, Cars, Aircrafts, and Actions, along with ImageNet that is considered as a generalist model or initial pre-existing model.

We compare again with \textit{Joint Training}, where we fine-tune the ImageNet pre-trained AlexNet jointly on the six tasks assuming all the data is available. We also compare with a setting with multiple fine-tuned models where the model with the maximum score is selected (\textit{Most confident model}).
For our Expert Gate, we follow the same regime as in the previous experiment. The most related task is always ImageNet. Based on our task relatedness threshold, LwF was selected for Actions, while Aircrafts and Cars were fine-tuned.
  \begin{figure*}[t]
\centering
\includegraphics[width=0.9\textwidth]{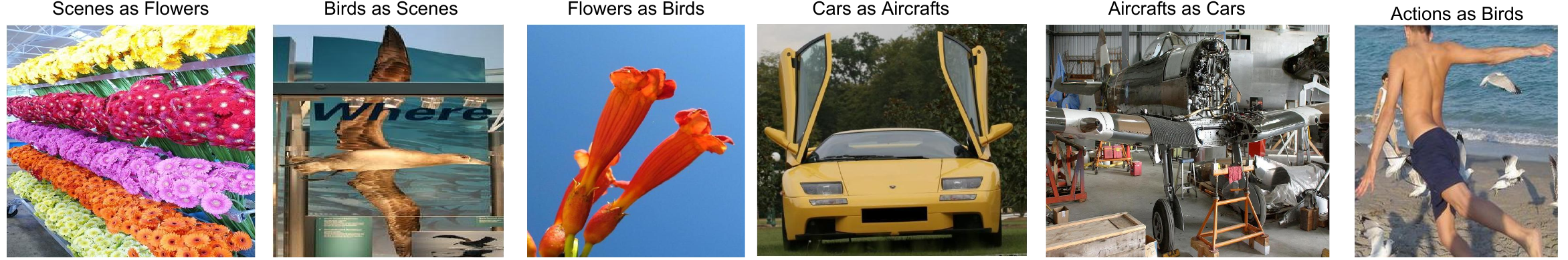}
  \caption{Examples of confusion cases made by our Expert Gate.}
  \label{fig:examples}
  \vspace*{-0.5cm} 
  \end{figure*}
Table \ref{tab:six} shows the results.

Even though the jointly trained model has been trained on  all the previous tasks data simultaneously, its average performance is inferior to our Expert Gate system.
This can be explained by the negative inductive bias where some tasks negatively affect others, as is the case for Scenes and Cars.

As we explained in the Introduction, deploying all  models and taking the max score (\textit{Most confident model}) is not an option: for many test samples the most confident model is not the correct one, resulting in poor performance. Additionally, with the size of each expert model around 220 MB and the size of each  autoencoder  around 28 MB, there is almost an order of magnitude difference in memory requirements.

\begin{figure}
\centering
\includegraphics[width=0.48\textwidth]{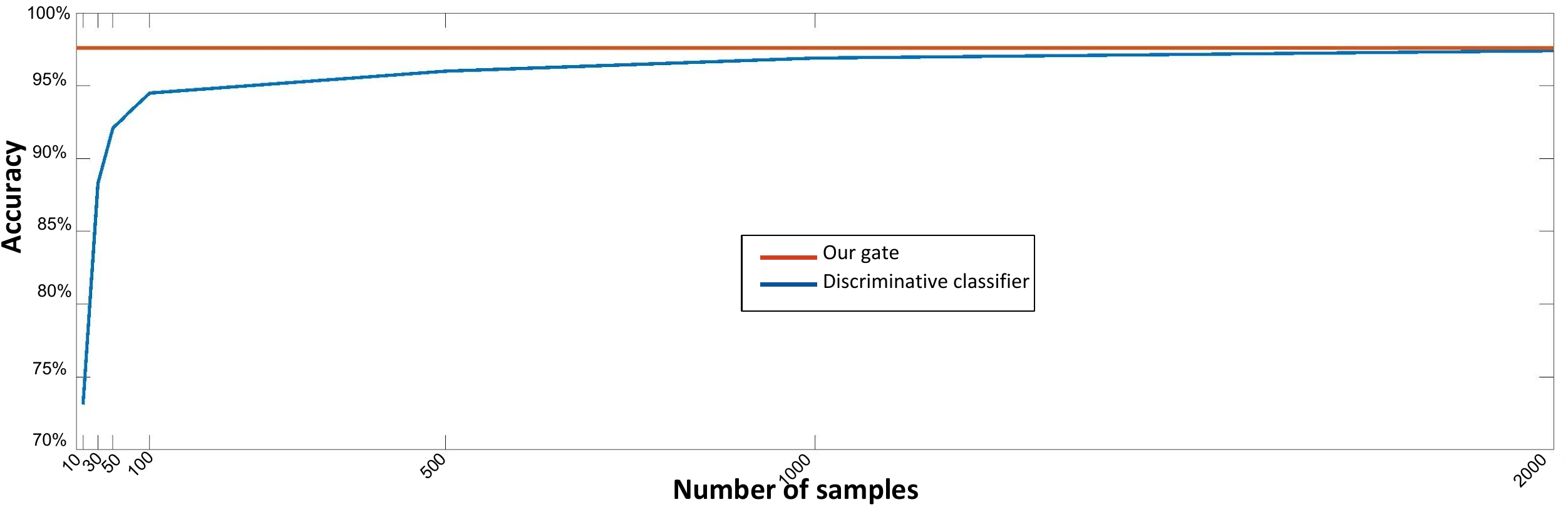}
  \caption{Comparison between our gate and the discriminative classifier with varying number of stored samples per task.}
   
  \label{fig:dis_class}
  \vspace*{-0.5cm} 
\end{figure}

{\bf Comparison with a discriminative classifier} Finally, we compare with a discriminative classifier trained to predict the task. For this classifier, we first  assume that all data from the previous tasks are stored, even though this is not in line with a lifelong learning setup. 
Thus, it serves as an upper bound.
For this classifier (\textit{Discriminative Task Classifier}) we use a neural net with one hidden layer composed of 100 neurons (same as our autoencoder code size). It takes as input the same data representation as our autoencoder gate and its output is the different tasks labels.
Table \ref{tab:gate} compares the performance of our gate on recognizing each task data to that of the discriminative classifier.
Further, we test the scenario of a discriminative classifier with the number of stored samples per task varying from 10-2000 (Figure \ref{fig:dis_class}). It approaches the accuracy of our gate with 2000 samples. Note that this is  $\frac{1}{2}$ to $\frac{1}{3}$ of the size of the used datasets. For larger datasets, an even higher number of samples would probably be needed to match performance.
In spite of not having access to any of the previous tasks data, our Expert Gate achieves similar performance to the discriminative classifier. In fact, our Expert Gate can be seen as a sequential classifier with new classes arriving one after another. This is one of the most important results from this paper:  {\em without ever having simultaneous access to the data of  different tasks, our Expert Gate based on autoencoders manages to assign test samples to the relevant tasks equally accurately as a discriminative classifier}.

Figure \ref{fig:examples} shows some of the few confusion cases for our Expert Gate.
For some test samples even humans have a hard time telling which expert should be activated. For example, Scenes images containing humans can also be classified as Actions. To deal with such cases, it may be preferable in some settings to allow more than one expert to be activated. This can be done by setting a threshold on the probabilities for the different tasks. We tested this scenario with a threshold of 0.1 and  observed 
$3.7\%$ of the test samples being analyzed by multiple expert models. Note that in this case we can only evaluate the label given by the corresponding task as we are missing the ground truth for the other possible tasks appearing in the image. This leads to an average accuracy of $68.2\%$, i.e. a further increase of $0.9\%$.

\subsection{Task Relatedness Analysis}
\label{sec:expres-related}
 \begin{figure}[t]
\centering
\includegraphics[width=0.5\textwidth]{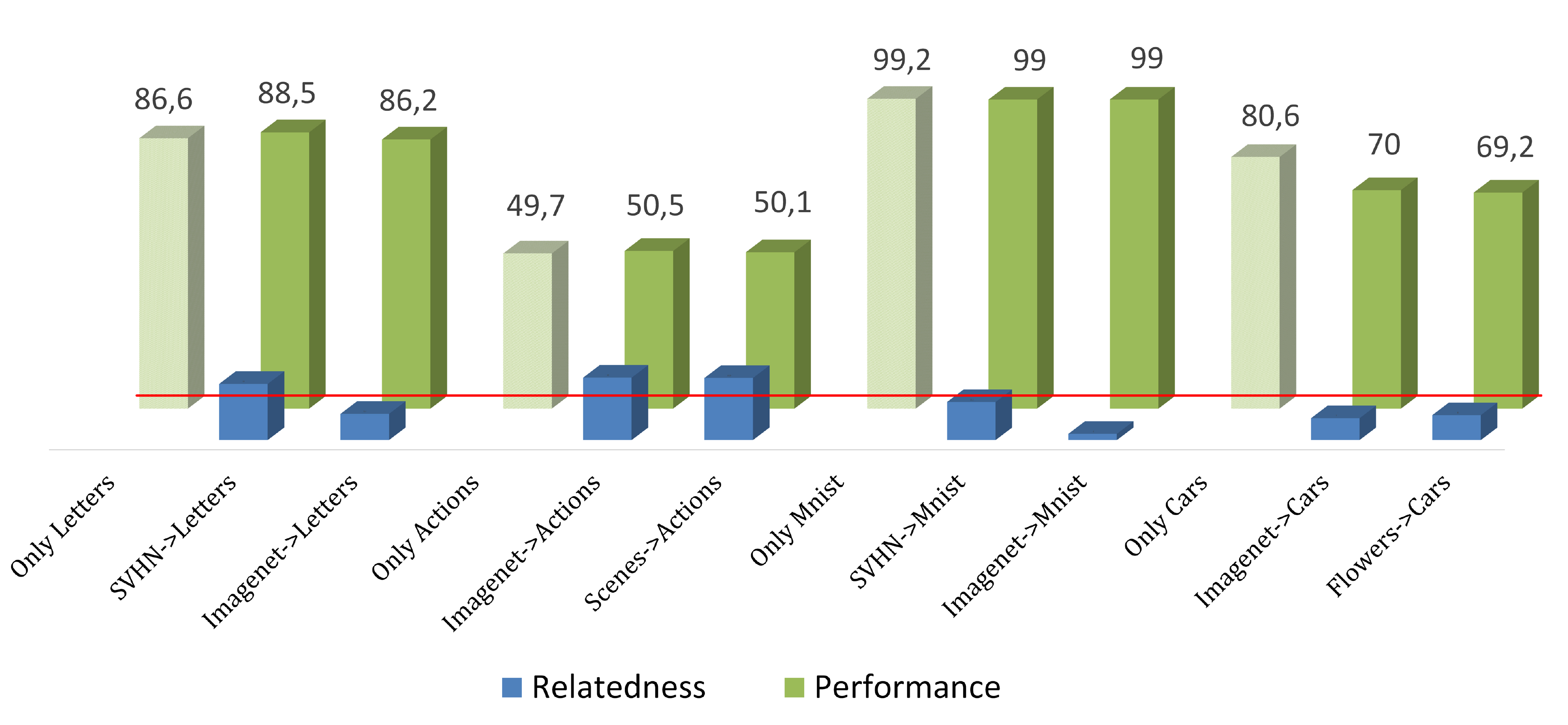}
  \caption{Relatedness analysis. The relatedness values are normalized for the sake of better visualization. The red line indicates our relatedness threshold value.}
  \label{fig:relatedness}
  \vspace*{-15pt} 
  \end{figure}
In the previous cases, the most related task was always Imagenet. This is due to the  similarity between the images of these different tasks and those of Imagenet. Also, the wide diversity of Imagenet classes enables it to cover a good range of these tasks. 
Does this mean that Imagenet should be the only task to transfer knowledge from, regardless of the current task nature? To answer this question, we add three more different tasks to our previous basket: the Google Street View House Numbers \textit{SVHN}~\cite{netzer2011reading} for digit recognition, the Chars74K dataset \cite{deCampos09} for character recognition in natural images (\textit{Letters}), and the \textit{Mnist} dataset \cite{lecun1998gradient} for handwritten digits.
For Chars74K, 
we use the English set and exclude the digits, considering only the characters.  From the previous set, we pick the two most related tasks, Actions and Scenes, and the two most unrelated tasks, Cars and Flowers. We focus on LwF \cite{li2016learning} as a method for knowledge transfer.  We also consider ImageNet as a possible source.  We consider the following knowledge transfer cases: Scenes $\rightarrow$ Actions, ImageNet $\rightarrow$ Actions, SVHN $\rightarrow$ Letters, ImageNet $\rightarrow$ Letters, SVHN $\rightarrow$ Mnist, ImageNet $\rightarrow$ Mnist, Flowers $\rightarrow$ Cars and Imagenet $\rightarrow$ Cars. 
Figure~\ref{fig:relatedness} shows the  performance of LwF compared to  fine-tuning  the tasks with pre-trained AlexNet (indicated by "Only X")  along with the degree of task relatedness. The red line indicates the threshold of $0.85$ task relatedness used in our previous experiments. 

In the case of a high score for task relatedness, the LwF  uses the knowledge from the previous task and improves  performance on the target task -- see e.g. (SVHN$\rightarrow$Letter, Scenes$\rightarrow$Actions, ImageNet$\rightarrow$Actions). When the tasks are less related, the method fails to improve and starts to degrade its performance, as in (Imagenet$\rightarrow$Letters, SVHN $\rightarrow$ Mnist). When the tasks are highly unrelated, LwF can even fail to reach a good performance for the new task, as in the case of (Imagenet$\rightarrow$ Cars, Flowers$\rightarrow$ Cars). This can be explained by the fact that each task is pushing the shared parameters in a different direction and thus the model fails to reach a good local minimum. We conclude that {\em our gate autoencoder succeeds to predict when a task could help another in the LwF framework and when it cannot}. 

\subsection{Video Prediction}
\label{sec:expres-video}
Next, we evaluate our Expert Gate for video prediction in the context of autonomous driving. 
We use a state of the art system for video prediction, the Dynamic Filter Network (DFN) \cite{de2016dynamic}. Given a sequence of 3 images, the task for the network is to predict the next 3 images. This is quite a structured task, where the task environment and training data affect the prediction results quite significantly. 
An autonomous vehicle that uses video prediction needs to be able to load the correct model for the current environment. It might not have all the data from the beginning, and so it becomes important to learn specialists for each type of environment, without the need for storing all the training data. Even when all data is available, joint training does not give the best results on each domain, as we show below. 

We show experiments conducted on three domains/tasks:
for \textit{Highway}, we use the data from DFN \cite{de2016dynamic}, with the same train/test split;  for \textit{Residential} data, we use the two longest sequences 
from the KITTI dataset \cite{geiger2013vision}; and for  \textit{City} data, we use the Stuttgart sequence from the CityScapes dataset \cite{Cordts2016Cityscapes}, i.e. the only sequence in that dataset with densely sampled frames. We use a 90/10 train/test split on both residential and city datasets. 
We train the 3 tasks using 3 different regimes: sequential training using a \textit{Single Fine-tuned Model}, \textit{Joint Training} and \textit{Expert Gate}. 
For video prediction, LwF does not seem applicable. In this experiment, we use the autoencoders only as  gating function. We do not use task relatedness.
Video prediction results are expressed as the average pixel-wise L1-distance between predicted
and ground truth images (lower is better), and shown in table \ref{tab:JntSeqTrainingVideoPred}. 

\begin{figure}[t]
\centering
\includegraphics[width=0.48\textwidth]{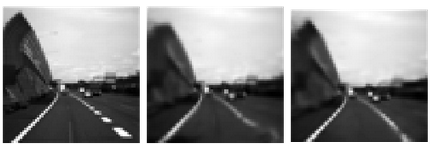}
\caption{Qualitative results for video prediction. From left to right: last ground truth image (in a sequence of 3); predicted image using sequential fine-tuning and using Expert Gate. Examining the lane markers, we see that 
Expert Gate is visually superior.}
\label{fig:qualResultsVideoPred1}
\vspace*{-0.5cm} 
\end{figure}

Similar trends are observed as for the image classification problem: sequential fine-tuning results in catastrophic forgetting, where a model fine-tuned on a new dataset deteriorates on the original dataset after fine-tuning. Joint training leads to better results on each domain, but requires all the data for training. Our Expert Gate system gives better results compared to both sequential and joint training. These numbers are supported by qualitative results as well (Figure \ref{fig:qualResultsVideoPred1}). Please refer to the supplementary materials for more figures.
These experiments show the potential of our Expert Gate system for video prediction tasks in autonomous driving applications.
\begin{table}
\footnotesize
\centering
\caption{ Video prediction results (average pixel L1 distance). For methods with * all the previous data needs to be available.}
\label{tab:JntSeqTrainingVideoPred}
\begin{tabular}{|l|c |c |c|c|}
    \hline
    Method &Highway&Residential&City &avg\\ \hline 
    Single Fine-tuned Model & 13.4&-&-&- \\ 
    					& 25.7&45.2 &-&- \\
    & 26.2& 50.0&17.3&31.1\\ \hline
     Joint Training* &14.0& 40.7&16.9& 23.8\\ \hline
     Expert Gate (ours) &\textbf{ 13.4}& \textbf{40.3}&\textbf{16.5}&\textbf{23.4}\\ \hline
\end{tabular}
\vspace*{-0.4cm} 
\end{table}
\section{Conclusions and Future Work}
\label{conclusions}
In the context of lifelong learning, most work has focused on how to exploit knowledge from previous tasks and transfer it to a new task. Little attention has gone to the related and equally important problem of how to select the proper (i.e. most relevant) model at test time. This is the topic we tackle in this paper.  
To the best of our knowledge, we are the first to propose a solution that does not require storing data from previous tasks. Surprisingly, Expert Gate's autoencoders can distinguish different tasks equally well as a discriminative classifier trained on all data. 
Moreover, they can be used to select the most related task and the most appropriate transfer method during training. 
Combined, this gives us a powerful method for lifelong learning, that outperforms not only the state-of-the-art 
but also joint training of all tasks simultaneously. \\
Our current system uses only the most related model for knowledge transfer.
As future work, we will explore the possibility of leveraging multiple related models for the training of new tasks -- for instance, by exploring new strategies for balancing the contribution of the different tasks by their relatedness degree rather than just varying the learning rates.
Also a mechanism to decide when to merge tasks with high relatedness degree rather than adding a new expert model, seems an interesting research direction.
\newline 
\small {\textbf{Acknowledgment:}
 The first author's PhD is funded by an FWO scholarship. We are grateful for support from KU Leuven GOA project CAMETRON.
The authors would like to thank Matthew B. Blaschko and Amal Rannen Triki for valuable discussions.
 }
{\small
\bibliographystyle{ieee}
\bibliography{egbib}

\begin{thebibliography}{10}\itemsep=-1pt

\bibitem{ahmed2016network}
K.~Ahmed, M.~H. Baig, and L.~Torresani.
\newblock Network of experts for large-scale image categorization.
\newblock {\em arXiv preprint arXiv:1604.06119}, 2016.

\bibitem{alain2014regularized}
G.~Alain and Y.~Bengio.
\newblock What regularized auto-encoders learn from the data-generating
  distribution.
\newblock {\em Journal of Machine Learning Research}, 15(1):3563--3593, 2014.

\bibitem{bengio2009learning}
Y.~Bengio et~al.
\newblock Learning deep architectures for ai.
\newblock {\em Foundations and trends{\textregistered} in Machine Learning},
  2(1):1--127, 2009.

\bibitem{bourlard1988auto}
H.~Bourlard and Y.~Kamp.
\newblock Auto-association by multilayer perceptrons and singular value
  decomposition.
\newblock {\em Biological cybernetics}, 59(4-5):291--294, 1988.

\bibitem{caruana1998multitask}
R.~Caruana.
\newblock Multitask learning.
\newblock In {\em Learning to learn}, pages 95--133. Springer, 1998.

\bibitem{Cordts2016Cityscapes}
M.~Cordts, M.~Omran, S.~Ramos, T.~Rehfeld, M.~Enzweiler, R.~Benenson,
  U.~Franke, S.~Roth, and B.~Schiele.
\newblock The cityscapes dataset for semantic urban scene understanding.
\newblock In {\em Proc. of the IEEE Conference on Computer Vision and Pattern
  Recognition (CVPR)}, 2016.

\bibitem{de2016dynamic}
B.~De~Brabandere, X.~Jia, T.~Tuytelaars, and L.~Van~Gool.
\newblock Dynamic filter networks.
\newblock {\em arXiv preprint arXiv:1605.09673}, 2016.

\bibitem{deCampos09}
T.~E. de~Campos, B.~R. Babu, and M.~Varma.
\newblock Character recognition in natural images.
\newblock In {\em Proceedings of the International Conference on Computer
  Vision Theory and Applications, Lisbon, Portugal}, February 2009.

\bibitem{pascal-voc-2012}
M.~Everingham, L.~Van~Gool, C.~K.~I. Williams, J.~Winn, and A.~Zisserman.
\newblock The {PASCAL} {V}isual {O}bject {C}lasses {C}hallenge 2012 {(VOC2012)}
  {R}esults.
\newblock
  http://www.pascal-network.org/challenges/VOC/voc2012/workshop/index.html.

\bibitem{geiger2013vision}
A.~Geiger, P.~Lenz, C.~Stiller, and R.~Urtasun.
\newblock Vision meets robotics: The kitti dataset.
\newblock {\em The International Journal of Robotics Research}, page
  0278364913491297, 2013.

\bibitem{Goodfellow-et-al-2016-Book}
I.~Goodfellow, Y.~Bengio, and A.~Courville.
\newblock Deep learning.
\newblock Book in preparation for MIT Press, 2016.

\bibitem{goodfellow2013empirical}
I.~J. Goodfellow, M.~Mirza, D.~Xiao, A.~Courville, and Y.~Bengio.
\newblock An empirical investigation of catastrophic forgetting in
  gradient-based neural networks.
\newblock {\em arXiv preprint arXiv:1312.6211}, 2013.

\bibitem{hinton2015distilling}
G.~Hinton, O.~Vinyals, and J.~Dean.
\newblock Distilling the knowledge in a neural network.
\newblock {\em arXiv preprint arXiv:1503.02531}, 2015.

\bibitem{hinton2006reducing}
G.~E. Hinton and R.~R. Salakhutdinov.
\newblock Reducing the dimensionality of data with neural networks.
\newblock {\em Science}, 313(5786):504--507, 2006.

\bibitem{jacob2009clustered}
L.~Jacob, J.-p. Vert, and F.~R. Bach.
\newblock Clustered multi-task learning: A convex formulation.
\newblock In {\em Advances in neural information processing systems}, pages
  745--752, 2009.

\bibitem{jacobs1991adaptive}
R.~A. Jacobs, M.~I. Jordan, S.~J. Nowlan, and G.~E. Hinton.
\newblock Adaptive mixtures of local experts.
\newblock {\em Neural computation}, 3(1):79--87, 1991.

\bibitem{kokkinos2016ubernet}
I.~Kokkinos.
\newblock Ubernet: Training auniversal'convolutional neural network for low-,
  mid-, and high-level vision using diverse datasets and limited memory.
\newblock {\em arXiv preprint arXiv:1609.02132}, 2016.

\bibitem{krause20133d}
J.~Krause, M.~Stark, J.~Deng, and L.~Fei-Fei.
\newblock 3d object representations for fine-grained categorization.
\newblock In {\em Proceedings of the IEEE International Conference on Computer
  Vision Workshops}, pages 554--561, 2013.

\bibitem{kumar2012learning}
A.~Kumar and H.~Daume~III.
\newblock Learning task grouping and overlap in multi-task learning.
\newblock {\em arXiv preprint arXiv:1206.6417}, 2012.

\bibitem{lecun1998gradient}
Y.~LeCun, L.~Bottou, Y.~Bengio, and P.~Haffner.
\newblock Gradient-based learning applied to document recognition.
\newblock {\em Proceedings of the IEEE}, 86(11):2278--2324, 1998.

\bibitem{li2016learning}
Z.~Li and D.~Hoiem.
\newblock Learning without forgetting.
\newblock In {\em European Conference on Computer Vision}, pages 614--629.
  Springer, 2016.

\bibitem{maji13fine-grained}
S.~Maji, J.~Kannala, E.~Rahtu, M.~Blaschko, and A.~Vedaldi.
\newblock Fine-grained visual classification of aircraft.
\newblock Technical report, 2013.

\bibitem{mante2013context}
V.~Mante, D.~Sussillo, K.~V. Shenoy, and W.~T. Newsome.
\newblock Context-dependent computation by recurrent dynamics in prefrontal
  cortex.
\newblock {\em Nature}, 503(7474):78--84, 2013.

\bibitem{marc2007unified}
Y.~Marc’Aurelio~Ranzato and L.~B. S. C.~Y. LeCun.
\newblock A unified energy-based framework for unsupervised learning.
\newblock In {\em Proc. Conference on AI and Statistics (AI-Stats)}, volume~24,
  2007.

\bibitem{mitchell1980need}
T.~M. Mitchell.
\newblock {\em The need for biases in learning generalizations}.
\newblock Department of Computer Science, Laboratory for Computer Science
  Research, Rutgers Univ. New Jersey, 1980.

\bibitem{netzer2011reading}
Y.~Netzer, T.~Wang, A.~Coates, A.~Bissacco, B.~Wu, and A.~Y. Ng.
\newblock Reading digits in natural images with unsupervised feature learning.
\newblock 2011.

\bibitem{nguyen2015deep}
A.~Nguyen, J.~Yosinski, and J.~Clune.
\newblock Deep neural networks are easily fooled: High confidence predictions
  for unrecognizable images.
\newblock In {\em 2015 IEEE Conference on Computer Vision and Pattern
  Recognition (CVPR)}, pages 427--436. IEEE, 2015.

\bibitem{Nilsback08}
M.-E. Nilsback and A.~Zisserman.
\newblock Automated flower classification over a large number of classes.
\newblock In {\em Proceedings of the Indian Conference on Computer Vision,
  Graphics and Image Processing}, Dec 2008.

\bibitem{quattoni2009recognizing}
A.~Quattoni and A.~Torralba.
\newblock Recognizing indoor scenes.
\newblock In {\em Computer Vision and Pattern Recognition, 2009. CVPR 2009.
  IEEE Conference on}, pages 413--420. IEEE, 2009.

\bibitem{ILSVRC15}
O.~Russakovsky, J.~Deng, H.~Su, J.~Krause, S.~Satheesh, S.~Ma, Z.~Huang,
  A.~Karpathy, A.~Khosla, M.~Bernstein, A.~C. Berg, and L.~Fei-Fei.
\newblock {ImageNet Large Scale Visual Recognition Challenge}.
\newblock {\em International Journal of Computer Vision (IJCV)},
  115(3):211--252, 2015.

\bibitem{rusu2016progressive}
A.~A. Rusu, N.~C. Rabinowitz, G.~Desjardins, H.~Soyer, J.~Kirkpatrick,
  K.~Kavukcuoglu, R.~Pascanu, and R.~Hadsell.
\newblock Progressive neural networks.
\newblock {\em arXiv preprint arXiv:1606.04671}, 2016.

\bibitem{silver2002task}
D.~L. Silver and R.~E. Mercer.
\newblock The task rehearsal method of life-long learning: Overcoming
  impoverished data.
\newblock In {\em Conference of the Canadian Society for Computational Studies
  of Intelligence}, pages 90--101. Springer, 2002.

\bibitem{silver2013lifelong}
D.~L. Silver, Q.~Yang, and L.~Li.
\newblock Lifelong machine learning systems: Beyond learning algorithms.
\newblock In {\em AAAI Spring Symposium: Lifelong Machine Learning}, pages
  49--55. Citeseer, 2013.

\bibitem{terekhov2015knowledge}
A.~V. Terekhov, G.~Montone, and J.~K. O’Regan.
\newblock Knowledge transfer in deep block-modular neural networks.
\newblock In {\em Conference on Biomimetic and Biohybrid Systems}, pages
  268--279. Springer, 2015.

\bibitem{thrun1998clustering}
S.~Thrun and J.~O’Sullivan.
\newblock Clustering learning tasks and the selective cross-task transfer of
  knowledge.
\newblock In {\em Learning to learn}, pages 235--257. Springer, 1998.

\bibitem{vedaldi2015matconvnet}
A.~Vedaldi and K.~Lenc.
\newblock Matconvnet: Convolutional neural networks for matlab.
\newblock In {\em Proceedings of the 23rd ACM international conference on
  Multimedia}, pages 689--692. ACM, 2015.

\bibitem{WelinderEtal2010}
P.~Welinder, S.~Branson, T.~Mita, C.~Wah, F.~Schroff, S.~Belongie, and
  P.~Perona.
\newblock {Caltech-UCSD Birds 200}.
\newblock Technical Report CNS-TR-2010-001, California Institute of Technology,
  2010.

\bibitem{xue2007multi}
Y.~Xue, X.~Liao, L.~Carin, and B.~Krishnapuram.
\newblock Multi-task learning for classification with dirichlet process priors.
\newblock {\em Journal of Machine Learning Research}, 8(Jan):35--63, 2007.

\bibitem{zeiler2013rectified}
M.~D. Zeiler, M.~Ranzato, R.~Monga, M.~Mao, K.~Yang, Q.~V. Le, P.~Nguyen,
  A.~Senior, V.~Vanhoucke, J.~Dean, et~al.
\newblock On rectified linear units for speech processing.
\newblock In {\em 2013 IEEE International Conference on Acoustics, Speech and
  Signal Processing}, pages 3517--3521. IEEE, 2013.

\end{thebibliography}
}

\end{document}